%% file: main.tex
\documentclass[runningheads]{style-eccv/llncs}

\usepackage{style-eccv/eccv}

\input{0-imports.tex}
\input{0-macros.tex}
\input{0-math.tex}
\input{0-space-saving.tex}

\title{Audio-Synchronized Visual Animation}

\author{
Lin Zhang\inst{1}\orcidlink{0000-0002-0955-6510} \and
Shentong Mo\inst{2}\orcidlink{0000-0003-3308-9585} \and
Yijing Zhang\inst{1}\orcidlink{0009-0008-2848-8649} \and
Pedro Morgado\inst{1}\orcidlink{0000-0002-0955-6510}
}

\authorrunning{Lin Zhang et al.}
\titlerunning{Audio-Synchronized Visual Animation}

\institute{
University of Wisconsin-Madison \and
Carnegie Mellon University
}

\begin{document}

\maketitle
\begin{center}
    \centering
    \captionsetup{type=figure}
    \includegraphics[width=\linewidth]{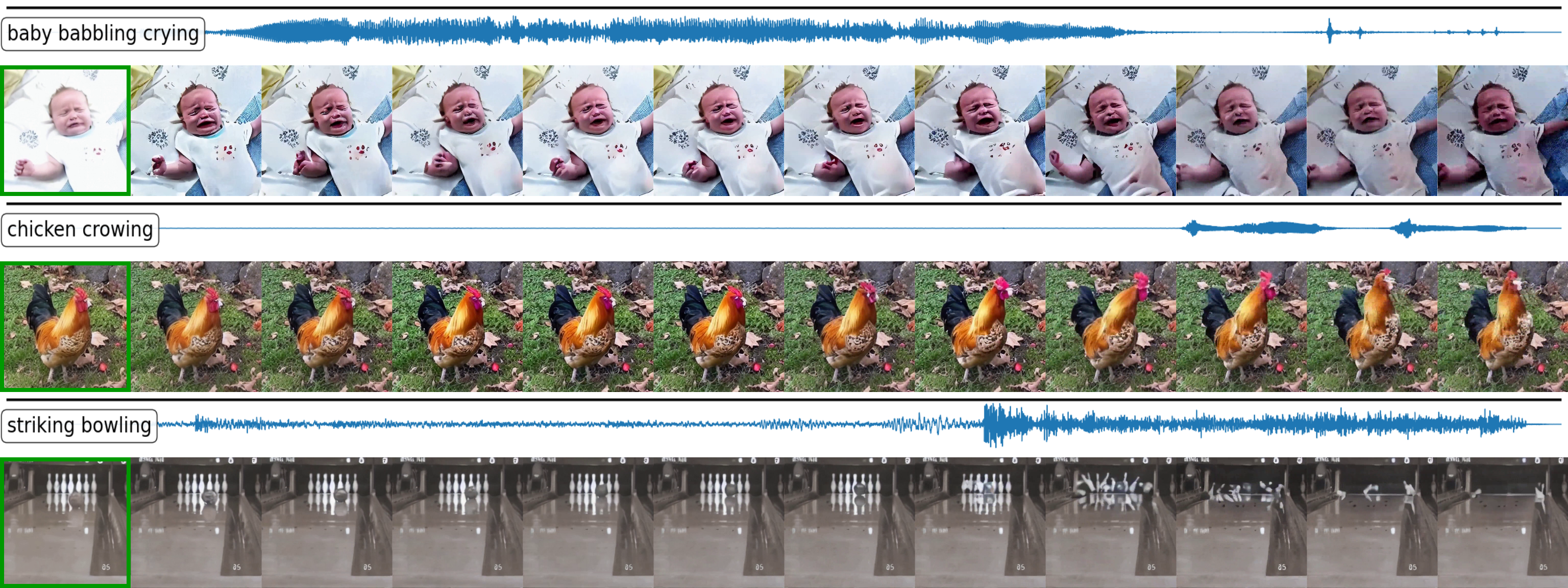}
    \captionof{figure}{Given an audio and an image~(\textcolor{ForestGreen}{green box}), we produce animations beyond image stylization with complex but natural dynamics, synchronized with audio at each frame. Results were produced by our \method model trained on the proposed \db. 
    Project webpage: \url{https://lzhangbj.github.io/projects/asva/asva.html}.}\label{fig:qualitative_view}
\end{center}%

\input{latex/0-abstract}

\input{latex/1-intro}
\input{latex/2-related-work}
\input{latex/3-dataset}
\input{latex/4-method}
\input{latex/5-experiments}
\input{latex/6-conclusion}

\bibliographystyle{style-eccv/splncs04}
\bibliography{refs}


\end{document}

%% file: 0-imports.tex
\usepackage{style-eccv/eccvabbrv}
\usepackage{graphicx}
\usepackage{booktabs}
\usepackage[accsupp]{axessibility}  

\usepackage{hyperref}
\usepackage{orcidlink}

\usepackage{ulem}

\usepackage{multirow}
\usepackage{makecell}

\usepackage{xparse}
\usepackage{pifont}

\usepackage{textcomp}  
\usepackage{scalerel}  
\def\fire{\scaleto[10pt]{\includegraphics{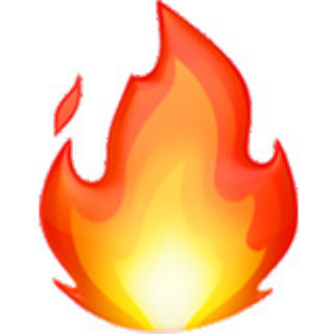}}{8pt}}

\usepackage{tabularx,colortbl}

\usepackage{enumitem}

%% file: 0-macros.tex
\newcommand{\cmark}{\ding{51}}%
\newcommand{\xmark}{\ding{55}}%

\newlength\savewidth
\newlength\thinwidth

\definecolor{Gray}{gray}{0.93}
\definecolor{ForestGreen}{rgb}{0.13, 0.75, 0.13}

\usepackage{xspace} 
\makeatletter
\DeclareRobustCommand\onedot{\futurelet\@let@token\@onedot}
\def\@onedot{\ifx\@let@token.\else.\null\fi\xspace}

\makeatother

\newcommand{\method}{{\selectfont{AVSyncD}}\xspace}
\newcommand{\task}{{\selectfont{ASVA}}\xspace}
\newcommand{\db}{{\selectfont{AVSync15}}\xspace}

%% file: 0-math.tex
\usepackage{amsmath,amsfonts,bm}

\def\1{\bm{1}}

\def\vtau{{\bm{\tau}}}

\def\va{{\bm{a}}}

\def\vv{{\bm{v}}}

\def\vx{{\bm{x}}}

\def\vz{{\bm{z}}}

\DeclareMathAlphabet{\mathsfit}{\encodingdefault}{\sfdefault}{m}{sl}
\SetMathAlphabet{\mathsfit}{bold}{\encodingdefault}{\sfdefault}{bx}{n}

\def\gD{{\mathcal{D}}}
\def\gE{{\mathcal{E}}}

\def\gL{{\mathcal{L}}}

\def\gN{{\mathcal{N}}}

\def\sE{{\mathbb{E}}}

%% file: 0-space-saving.tex
\newcommand{\newparagraph}[1]{\vspace{0.1em} \noindent {\bf #1}}



%% file: latex/0-abstract.tex
\begin{abstract}
Current visual generation methods can produce high-quality videos guided by text prompts. However, effectively controlling object dynamics remains a challenge. This work explores audio as a cue to generate temporally synchronized image animations. We introduce Audio-Synchronized Visual Animation (\task), a task that aims to animate a static image of an object with motions temporally guided by audio clips. To this end, we present \db, a dataset curated from VGGSound with videos featuring synchronized audio-visual events across 15 categories. We also present a diffusion model, \method, capable of generating audio-guided animations. 
Extensive evaluations validate \db as a reliable benchmark for synchronized generation and demonstrate our model's superior performance. We further explore \method's potential in a variety of audio-synchronized generation tasks, from generating full videos without a base image to controlling object motions with various sounds. We hope our established benchmark can open new avenues for controllable visual generation.

\keywords{Audio \and Video \and Synchronization \and Generation}

\end{abstract}

%% file: latex/1-intro.tex
\section{Introduction}\label{sec:intro}

Generative modeling has witnessed remarkable progress recently, largely due to the emergence of diffusion models~\cite{ho2020ddpm,song2022ddim,robin2022ldm}. Conditional generation and, in particular, text-to-image generation~\cite{ramesh2022dalle2,robin2022ldm}, has been the focal point given its application potential and availability of high-quality datasets~\cite{schuhmann2022laion5b,Bain21webvid10m}. This success has also led to revived interest in video generation, such as text-to-video~\cite{vid2vid-zero,text2video-zero,singer2022makeavideo,chen2023videocrafter1}. While text guidance has been thoroughly investigated, the unique advantages of audio-visual synchrony for video generation remain underexplored. Unlike text, which provides control over global semantics, audio offers both semantic control on videos and precise control at each moment in time.

Most existing works on audio-to-visual generation are however either limited to semantic control~\cite{lee2022soundguided,jeong2023tpos,sung2023soundtovisual,tang2023codi}, or constrained on singular scenarios such as human talking face~\cite{lipreading,zhou2019talking,park2022synctalkface,ye2023geneface,Zhou2021Pose,ng2024audio2photoreal,outoftime,Chung2019perfectmatch}. The former focuses on ambient audio datasets lacking synchronization cues with environmental sounds~(ranining, fire crakcling, wind)~\cite{lee2022soundguided} where shifting the audio temporally does not lead to major changes in visual content. The input audio in such cases thus substitutes text to provide only global semantics. The latter, talking faces, while synchronized, are extremely limited in generation diversity and control.


To bridge this gap, we introduce Audio-Synchronized Visual Animation, \task, a task which aims to animate objects depicted in natural static images into a video, with clear \emph{motion dynamics} that are \emph{semantically aligned} and \emph{temporally synchronized} with the input audio.~\task{} requires a sophisticated understanding of the audio's temporal structure, as well as of how objects move in synchrony with sound. Prior attempts on visual generation guided by \textit{diverse sounds} fall short in generated visual quality~\cite{lee2023aadiff} and accurate synchronization control~\cite{yariv2023tempotoken,jeong2023tpos} due to two challenges: (1) the lack of high-quality training data and benchmarks for learning audio-synchronized visual dynamics; (2) the development of effective methods capable of generating highly synchronized video motions. Successfully addressing these challenges will expand the scope of current video generation methods to enable more fine-grained semantic and temporal control over the generation process via synchronized audio conditioning.

We address the first challenge by constructing a high-quality diverse dataset with strong correlations between audio and object motions at each moment in time. In an ideal dataset, sound sources should be easily identifiable in the scene. Every visual motion in the video should highly correlate with the audio semantically and temporally, and vice versa. Moreover, the visual content should be of high quality for generation. However, existing audio-visual datasets either are too noisy~\cite{audioset,chen2020vggsound}, containing a large number of unassociated audio-visual pairs~\cite{morgado2021_robust_xid}, or predominantly featuring ambient sound categories that lack meaningful synchronized object dynamics~\cite{lee2022soundguided,aist-dance-db}. We thus curate a high-quality dataset from VGGSound~\cite{chen2020vggsound} by deploying an efficient two-step curation pipeline. In the first step, we use a variety of signal processing techniques and foundation models to automatically filter out videos with poor semantic alignment or temporal synchronization, as well as those depicting static scenes or with too fast camera motions. Then, to ensure the highest possible quality as a benchmark, we narrow down the dataset to sound categories with strong audio-visual synchronization cues, and manually verify the quality of each video. We end up obtaining \db  with 15 dynamic sound classes, each with 100 examples rich and accurate in semantics, object dynamics, and audio-visual synchronization. An overview of \db classes is in \cref{fig:avsync_gallery,fig:vggsound_dataset_compare} and its comparison with prior audio-video generation datasets is in Suppl. Sec. 5.1.

The second challenge pertains to the generation of audio-synchronized motions, which requires a detailed understanding of audio-visual correlations and object dynamics. Take, for instance, a video featuring a dog. To generate realistic video, the model is expected to not only synchronize the dog's mouth with the barking sound, but also accurately depict subtleties in the dog's head pose before and after barking. Furthermore, in a more challenging scenario, the model should discern which object to animate to preserve semantic consistency depending on the input sound. However, existing audio-conditioned visual generation frameworks~\cite{girdhar2023imagebind,sung2023soundtovisual} primarily focus on semantic control, often encoding the audio into a single global semantic feature and thereby neglecting the audio's temporal domain. Even recent attempts at audio-synchronized video generation~\cite{lee2023aadiff,jeong2023tpos,yariv2023tempotoken} have not fully realized the potential of audio for fine-grained temporal control, as they either rely on crude audio representations such as audio amplitude~\cite{lee2023aadiff}, learn from weakly synchronized~\cite{lee2022soundguided}, noisy datasets~\cite{chen2020vggsound}, or ignore object dynamics in the generation process~\cite{lee2023aadiff,lee2022soundguided,jeong2023tpos,yariv2023tempotoken}.
To this end, we introduce Audio-Video Synchronized Diffusion~(\method), a framework improving a pre-trained image latent diffusion model~\cite{robin2022ldm} for enhanced audio guidance and motion generation. We employ the pre-trained ImageBind~\cite{girdhar2023imagebind} encoder to encode audio into time-aware semantic tokens, then fuse them into each frame's latent features. This allows for precise audio guidance on video semantics and synchronization. To capture complex video motions, we add temporal attention layers to the diffusion model. Finally, to ensure faithful animation of the input image, we incorporate temporal convolutions and attention layers that always reference the input image, i.e., first-frame lookups.

With the carefully designed dataset and architecture, we are able to train a model specialized for \task{} and produce animations with more realistic and audio-synchronized contents than prior works~(\cref{fig:qualitative_view,fig:benchmark}). We provide thorough experiments to validate the effectiveness of \db{} and \method{} and demonstrate how to deploy \method{} for controllable generation, including amplifying audio guidance and semantic-aware object animation~(\cref{sec:ablation,sec:application}).

%% file: latex/2-related-work.tex
\section{Related Work}\label{sec:related-work}

\subsection{Controllable Visual Generation}
Many conditional visual generation models based on diffusion process~\cite{ho2020ddpm,song2022ddim} have emerged recently. Benefiting from more efficient architecture, large-scale training data~\cite{schuhmann2022laion5b}, and aligned semantic space~\cite{radford2021CLIP}, Latent Diffusion Model (LDM)~\cite{robin2022ldm} has achieved great success to generate realistic images conditioned on text. This inspired researchers to explore various diffusion-based visual generation tasks, such as text-to-video~\cite{text2video-zero,blattmann2023videoldm,vid2vid-zero,wu2023lamp,chen2023videocrafter1,wang2023modelscope}, audio-to-image~\cite{girdhar2023imagebind,sung2023soundtovisual}, and audio-to-video~\cite{tang2023codi,jeong2023tpos,yariv2023tempotoken,lee2023aadiff}. The architectures can be training-free~\cite{text2video-zero,vid2vid-zero,lee2023aadiff}, fully-trained~\cite{chen2023videocrafter1,wang2023modelscope}, or trained partially, which augments a pre-trained LDM by carefully adding some trainable layers~\cite{blattmann2023videoldm,jeong2023tpos,wu2023lamp}. Extensive works have also attempted to control the semantics of the generated content~\cite{mokady2022null,hertz2022prompt,zhang2023adding,li2023gligen}, while how to apply control in the temporal dimension remains under-explored.

In this work, we developed an image animation model \method{} to control generation \textit{semantically} and \textit{temporally} guided by audio. \method{} augments pre-trained StableDiffusion~\cite{robin2022ldm} with trainable temporal layers and audio conditioning mechanism, preserving training efficiency and generalizing well.

\subsection{Audio-to-Video Generation}
Traditionally, audio has been used as a temporal cue for talking face generation~\cite{zhou2019talking,park2022synctalkface,ye2023geneface,Zhou2021Pose,ng2024audio2photoreal}, where face and lip actions should be synchronized with audio at each frame. Many works also rely on complex inputs such as 3D meshes and human poses~\cite{ye2023geneface,ng2024audio2photoreal}. Although accurately synchronized, this line of research is extremely limited in scenarios and cannot generate videos for diverse audios. 

A series of works attempted to expand the class of audio by encoding sound into a global semantic condition for video generation~\cite{sung2023soundtovisual,girdhar2023imagebind,tang2023codi}, however often overlooked the temporal aspect inherent in audio. Some recent works, although divided audio features into time-aware segments as inputs~\cite{lee2022soundguided,ruan2022mmdiffusion}, failed to achieve promising visual quality or synchronize video motions with audios. AADiff~\cite{lee2023aadiff} is a training-free method re-weighting the text-image cross-attention map in LDM using audio amplitude at each frame, however can only control styles of each frame. TPoS~\cite{jeong2023tpos} learns segmented audio features aligned with CLIP~\cite{radford2021CLIP} using sophisticated modules and training losses, and feeds them into a pre-trained text-to-image model~\cite{robin2022ldm} for video generation. TempoToken~\cite{yariv2023tempotoken} also learns segmented audio features with a pre-trained audio encoder BEATs~\cite{Chen2022beats}, and fuses them into a pre-trained text-to-video model~\cite{wang2023modelscope}. However, primarily focused on monotonous sound classes in Landscapes~\cite{lee2022soundguided} or noisy audio-visual data in VGGSound~\cite{chen2020vggsound}, these methods are limited to generating video semantics without capturing the natural and synchronized dynamics of video content. Frozen generation architectures also prevent them from generating natural motions.

To address these limitations, we introduce \db, a high-quality dataset specifically designed for \task. \db{} stands out from previous efforts by focusing on synchronization cues between audio and visual dynamics, allowing for generating motions beyond mere visual effects. Once trained on \db, our \method{} can generalize to many applications to control video motions guided by audios, on which previous methods performed poorly.

%% file: latex/3-dataset.tex
\section{Audio-Synchronized Visual Animation}\label{sec:dataset}

Formally, the Audio-Synchronized Visual Animation (\task) task can be posed as follows. Given an audio clip $\va$ of length $T$ seconds and an image $\vx_1$, the goal is to generate the future video sequence $\vx_2,\ldots,\vx_{rT}$ (or $\vx_{2:rT}$ for short), where $r$ is the desired frame rate. Despite the simple formulation, this is a challenging task as the generated video sequence should (1) be of high visual quality, (2) be semantically aligned with the image $\vx_1$ and audio $\va$, (3) exhibit temporal coherence and (4) natural object motions temporally synchronized with the audio $\va$.
To facilitate research in \task, we introduce a new benchmark that includes a curated high-quality dataset and a suite of evaluation metrics designed to capture the various components of audio-synchronized generation.

\subsection{\db: A High-Quality Dataset for Audio Synchronized Video Generation}
Existing large-scale audio-visual datasets like VGGSound~\cite{chen2020vggsound,chen2021vggsoundsync,sparse2022iashin} and AudioSet~\cite{audioset} often contain amateur videos from platforms like YouTube. These videos, while diverse, can pose challenges for audio-synchronized video generation tasks due to rapid scene changes, camera motion, noisy audio, or out-of-frame sound sources. Prior work~\cite{li2021aist++,lee2022soundguided} has addressed this by focusing on simpler videos, such as those depicting fire crackling or weather patterns. However, such videos often lack strong synchronization cues between audio and visual motion, making them unsuitable for \task. To facilitate research in this area, we introduce a high-quality dataset specifically designed for audio-synchronized video generation, ensuring a close synchronization between audio and visuals. 
More specifically, our selection criteria to create the dataset were: (1) \textit{High Correlation}: Significant visual changes should be closely associated with audio at each timestamp, and vice versa. (2) \textit{Dynamic Content}: We included content rich in temporal changes, excluding ambient or monotonous classes. (3) \textit{Quality and Relevance}: Both video and audio needed to be clean, stable, and semantically aligned.

\input{figures/avsync_gallery_and_vggsound_compare}

\newparagraph{Preliminary Curation}
We start from VGGSound~\cite{chen2020vggsound}, a large-scale dataset with 309 diverse audio classes. Similar to VGGSoundSync~\cite{chen2021vggsoundsync}, we first narrow down to 149 classes with potentially clear audio-visual synchronized events, removing ambient classes, which is referred as VGGSS. We then deploy a sequence of automatic cleaning steps and a final manual selection step to identify appropriate videos. The procedures are summarized below and detailed in Suppl. Sec. 5.2.

\newparagraph{Automatic Curation}
We first use PySceneDetect~\cite{pyscenedetect} to cut videos with sharp scene changes into different scenes, which are still likely to contain low-quality short clips. To maximize usage, we split each scene into 3-second clips with 0.5-second strides, and discard unsuitable clips based on the following metrics:
\begin{itemize}[label={}, leftmargin=0pt]
    \item \uline{\textit{Raw Pixel Difference}}: We calculate average pixel distances between consecutive frames and remove clips with both small and large values, likely depicting either static or videos with excessive motion.
    \item \uline{\textit{Image Semantics Difference}}: Complementing the raw pixel analysis above, we also compute distances on CLIP~\cite{radford2021CLIP} image features, removing videos with small semantics changes such as zoom in/out or large semantic content transitions.
    \item \uline{\textit{Audio Waveform Amplitude}}: We exclude clips whose maximum audio waveform amplitude is low, indicating weak audio cues.
    \item \uline{\textit{Semantic Alignment}}: We compute the average image-audio~(IA) and image-text~(IT) alignment scores~\cite{radford2021CLIP} in a video using ImageBind~\cite{girdhar2023imagebind}, removing clips with low scores to ensure cross-modal semantic alignment.
    \item \uline{\textit{Audio-Video Synchronization}}: To measure audio-visual synchronization, we follow VGGSoundSync~\cite{chen2021vggsoundsync} to contrastively train an audio-visual synchronization classifier on VGGSS, ending up with a comparable $40.85\%$ test accuracy.  The model outputs an unbounded av-sync score $\phi_{\va_i, \vv_j}$ for an audio-video pair $(\va_i, \vv_j)$. During training, we compute $\phi$ for a synchronized pair $(\va_i, \vv_i)$ and its temporally-shifted pairs from the same instance. Contrastive loss is then applied to these shifted pairs to maximize the synchronization probability:
    \input{equations/sync_prob}
    to distinguish the synchronized pair from shifted ones. We use $P_{\text{Sync}}$ as a synchronization indicator to remove low-scoring clips. When computing $P_{\text{sync}}$, we discard the temperature parameter $\tau$ used to improve training efficacy. We detail the synchronization classifier and $P_{\text{sync}}$ in Suppl. Sec. 1 and Sec. 2.1, respectively.
\end{itemize}
We empirically determine metrics' thresholds by prioritizing quality, acknowledging that some acceptable clips might be discarded. After automatic curation, we merge all continuous 3-second clips from each video and remove categories with less than 100 examples to avoid class imbalance, resulting in AVSync-AC (AVSync w/ Automatic Curation) with 76 categories and 39,902 examples.

\newparagraph{Manual Curation}
We further select 15 diverse categories with clear audio-visual motion cues from AVSync-AC for manual refinement. The categories range from animals and human actions to triggered tools and musical instruments. Manual curation once again seeks to identify appropriate videos for \task{} with the criteria above: high correlation, dynamic content, quality and relevance. 

\newparagraph{Dataset Summary}
The final dataset, \db, contains 90 training and 10 testing videos per category, each 2$\sim$10 seconds long. We provide an overview of \db{} in~\cref{fig:avsync_gallery}. To validate our curation pipeline, we randomly sample three 1500-video splits on the selected 15 categories from VGGSS and AVSync-AC, and quantitatively compare them with \db{} in~\cref{fig:vggsound_dataset_compare} and \cref{tab:dataset_ablation}. We also compare \db{} with other audio-visual datasets in Suppl. Sec. 5.1.


\def\PAlign{{P_{\textit{Align}}}}
\def\PRelSync{P_{\textit{RelSync}}}
\def\PSyncGAlign{P_{\textit{Sync}|\textit{Align}}}
\def\PAlignSync{{P_{\textit{Sync}\&\textit{Align}}}}

\subsection{Evaluation Metrics}\label{section:dataset_metric}
\task{} is a multi-faceted generation task, necessitating high quality at both image and video level. At the image level, we follow previous conventions~\cite{brock2019biggan,dhariwal2021diffusionbeatgan} to use (1) Fréchet Inception Distance (\textbf{FID})~\cite{fid} to measure the quality of individual frames; (2) \textbf{IA}~\cite{girdhar2023imagebind}/\textbf{IT}~\cite{xu2021videoclip} to measure image-audio/image-text semantics alignment on CLIP~\cite{radford2021CLIP} space. At the video level, we use Fréchet Video Distance (\textbf{FVD})~\cite{fvd} to assess video quality. To measure audio-video synchronization, we compute the following metrics with the trained synchronization classifier:

\newparagraph{RelSync} During testing, we use the ground truth audio-visual pair ($\va, \vv$) as a reference to measure synchronization of the generated video $\boldsymbol{\hat{v}}$ as follows:
\input{equations/relsync}
Note that while this reference-based metric normalizes the score by the synchronization of the reference pair ($\va, \vv$), the metric can still be sensitive to the quality of the reference pair. In fact, evaluating synchronization on a dataset where even ground-truth audios and videos are ambiguously synchronized is less informative of the model capabilities, e.g., Landscapes~\cite{lee2022soundguided}.

\newparagraph{AlignSync} The synchronization classifier is only trained on semantically-aligned and temporally-shifted audio-video pairs sampled from the same instance (See Suppl. Sec. 1.2). RelSync is thus implicitly conditioned on semantics alignment, i.e., P(Sync|Align). To jointly measure semantics alignment and synchronization, we first approximate $P_{\text{Align}}$ similarly as RelSync:
\input{equations/align_prob}
where $\vv_1$ is the first frame input for animation, and $b$ is the number of generated frames. The generated first frame $\boldsymbol{\hat{v}}_1$ is eliminated because it is often a replicate of input $\vv_1$. By multiplying $P_{\text{Align}}$ with RelSync, we obtain the joint score:
\input{equations/alignsync}
which jointly measures semantic alignment and temporal synchronization between $\va$ and $\boldsymbol{\hat{v}}$.


These automated metrics are not always aligned with human preference. We therefore conduct a user study detailed in~\cref{sec:main_results}, asking human raters to compare videos generated by multiple models and select the best one.

%% file: figures/avsync_gallery_and_vggsound_compare.tex
\begin{figure}[t!]
    \centering
     \begin{subfigure}[b]{0.39\linewidth}
         \centering
         \includegraphics[width=\linewidth]{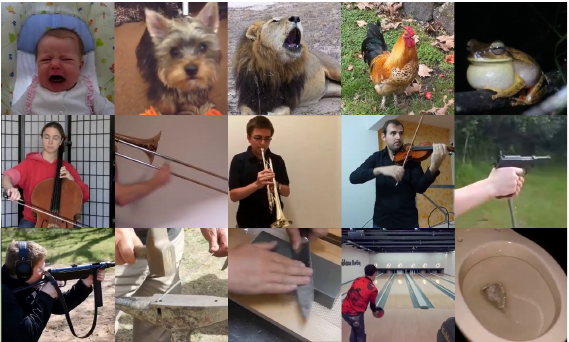}
         \caption{}
         \label{fig:avsync_gallery}
     \end{subfigure}
     \hfill
    \centering
    \begin{subfigure}[b]{0.58\linewidth}
        \includegraphics[width=\linewidth]{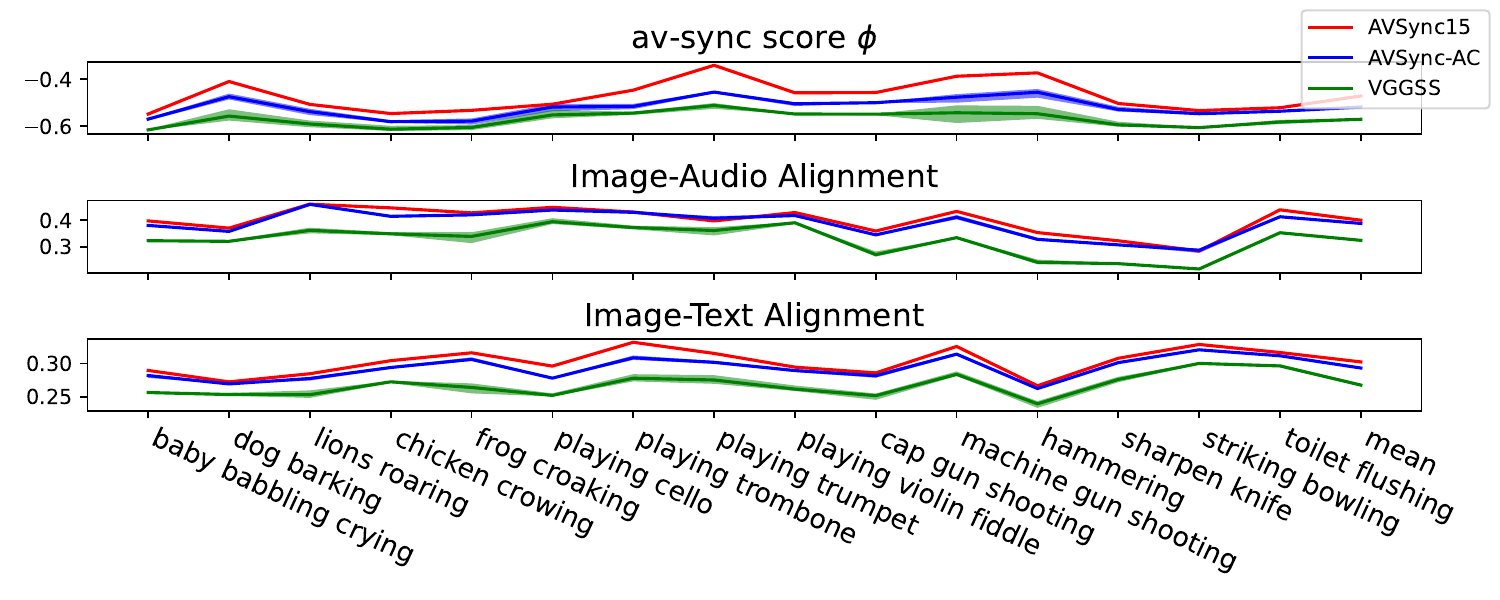}
        \caption{}
        \label{fig:vggsound_dataset_compare}
    \end{subfigure}
    \caption{(a): Overview of 15 categories in \db. Categories are listed below x-axis on the right plot. 
    (b): Categoriy-wise averages of av-sync score $\phi$, IA, and IT on \db and equivalently sized subsets of VGGSS and AVSync-AC. Error bars for VGGSS and AVSync-AC are obtained from 3 random splits.}
\end{figure}

%% file: equations/sync_prob.tex
\begin{equation}
    P_{\text{Sync}}(\va_i, \vv_i) = \frac{\exp(\phi_{\va_i, \vv_i} / \tau)}{\sum_{j}\exp(\phi_{\va_i, \vv_j} / \tau)}
    \label{equation:sync_prob}
\end{equation}

%% file: equations/relsync.tex
\begin{equation}
    \text{RelSync}(\va, \boldsymbol{\hat{v}}, \vv) = \frac{\exp(\phi_{\va, \boldsymbol{\hat{v}}})}{\exp(\phi_{\va, \vv}) + \exp(\phi_{\va, \boldsymbol{\hat{v}}})}
\label{equation:suppl_relsync}
\end{equation}

%% file: equations/align_prob.tex
\begin{equation}
        P_{\text{Align}}(\va, \boldsymbol{\hat{v}}, \vv_1) = \frac{1}{b-1} \sum_{i=2...b} \frac{\exp(\text{IA}_{\va, \boldsymbol{\hat{v}}_i})}{\exp(\text{IA}_{\va, \boldsymbol{\hat{v}}_i}) + \exp(\text{IA}_{\va, \vv_1})}
        \label{equation:align_prob}
\end{equation}

%% file: equations/alignsync.tex
\begin{equation}
    \text{AlignSync}(\va, \boldsymbol{\hat{v}}, \vv) = P_\text{Align}(\va, \boldsymbol{\hat{v}}, \vv_1) \cdot \text{RelSync}(\va, \boldsymbol{\hat{v}}, \vv)
    \label{equation:alignsync}
\end{equation}

%% file: latex/4-method.tex
\section{Audio-Video Synchronized Diffusion}\label{sec:method}

\subsection{Preliminary: Text-to-Image Latent Diffusion}

Text-to-image latent diffusion models~(LDMs~\cite{robin2022ldm}) encode images $\vx$ into a lower-dimensional latent space $\vz=\gE(\vx)$ using a pre-trained perceptual auto-encoder $\gE(\cdot)$, and learn the conditional distribution $p(\vz|\vtau)$ given a CLIP-encoded text prompt $\vtau$. It gradually denoises latents $\vz^k$, obtained by corrupting the image latent $\vz$ with Gaussian noise $\epsilon$, over $k$ time steps. A denoising UNet $\epsilon_\theta(\cdot)$ parameterized by $\theta$ is deployed to estimate the added noise $\epsilon$ by minimizing
\begin{equation}
    \gL_{\textit{LDM}} = \sE_{\vz, \epsilon\sim \gN(0, 1), k}\left[\| \epsilon - \epsilon_\theta(\vz^k, k, \vtau) \|_2^2\right]
    \label{equation:diffusion_loss}
\end{equation}
During inference, LDMs start from a random Gaussian noise map $\vz^K$, and iterate over $K$ reverse diffusion steps~\cite{ho2020ddpm,song2022ddim} to denoise the latents $\vz^{k-1} = \vz^{k}-\epsilon_\theta(\vz^k, k, \vtau)$, until $\vz^0$ is obtained. LDMs then decode the latent into an image $\vx^0=\gD(\vz^0)$ using the pre-trained decoder $\gD(\cdot)$. For simplicity, we refer to the images by their latents $\vz$ rather than $\vx$ throughout the rest of the paper.

\subsection{Proposed Architecture}
In this work, we extend the capabilities of LDMs for \task, by focusing on learning video dynamics and temporal synchronization, unlike existing approaches~\cite{girdhar2023imagebind,sung2023soundtovisual,tang2023codi} which primarily use audio to control global semantics. 
We propose the Audio-Video Synchronized Diffusion model~(\method), which builds upon a pre-trained image LDM and integrates synchronized audio control and temporal layers for improved video consistency. The major component is a UNet $\epsilon_\theta$ aiming to denoise a video instead of an image.
The UNet is trained on synchronized audio-video pairs, with the first frame $\vz_1$, the corresponding audio $\va$, and the CLIP encoded audio category name $\vtau$ as input conditions. The LDM denoising objective in \cref{equation:diffusion_loss} is applied to the remaining frames $\vz_{2:rT}$ to be predicted. The architecture of \method{} is shown in \cref{fig:arch_overview} and discussed below. Further details, for example, highlighting the different attention modules used in the architecture are described in Suppl. Sec. 3.2.

\input{figures/arch_overview}

\newparagraph{Initial-frame Conditioning}
To condition LDM on an input image, we feed its latent $\vz_1$ without noise into the UNet at every diffusion timestep $k$. For all subsequent frames, we adhere to the original LDM, using independently sampled noisy latents $\vz_{2:rT}^k$ as inputs and predicting the added noise $\epsilon^k$.

\newparagraph{Text Conditioning} We retain the text cross-attention layers in the original LDMs~\cite{robin2022ldm} (without finetuning) to condition the model on the audio category. However, due to the limited text diversity in training, class conditioning does not bring significant gains (see Suppl. Sec. 6.5).

\newparagraph{Audio Conditioning}
To facilitate audio-synchronized generation, we condition the generation on ImageBind audio embeddings~\cite{girdhar2023imagebind}. ImageBind computes an audio classification token, $\va^g$, representing global semantics, and local patch tokens, $\va_{f,t}$, across $T_a$ timestamps. The original ImageBind only uses $\va^g$ for contrastive learning and discards $\va_{f,t}$. However, we found these frozen local tokens as efficient synchronization cues. We split the patch tokens temporally into $rT$ segments, corresponding to the same timestamps as the frames $\vz_{1:rT}$, and append the global token to each. Each frame $\vz_t$ then learns both semantics and synchronization guidance from its audio segment $\va_t$ via cross-attention~\cite{vaswani2023attention}. In Suppl. Sec. 6.7, we compared ImageBind with CLAP and BEATs as encoders.

\newparagraph{Spatial Convolution}
The original LDM's pre-trained spatial convolutional layers were frozen and used without modification.

\newparagraph{First-frame Temporal Convolution} 
Each spatial convolution block was augmented with a 1D temporal convolution layer (kernel size 3) to capture temporal dependencies. To better adhere to the starting image, $\vz_1$, we adjusted the receptive field at frame $t$ to include frames $(1, t-1, t)$ as opposed to $(t-1,t,t+1)$. These temporal convolutions with \emph{first-frame lookup} were applied to three components in the UNet, namely the input/output conv layers and all ResNet convs.

\newparagraph{First-frame Spatial Attention}
We also leveraged the pre-trained LDM's spatial self-attentions. Following~\cite{text2video-zero}, we modified the frozen spatial attention layers to cross-attend to the first frame rather than self-attend to the current frame, by computing the key-value pairs from the first frame and the queries from the current one.

\newparagraph{Temporal Attention}
To effectively model long-range visual dependencies, we incorporated temporal attention layers~\cite{blattmann2023videoldm}. Each frame index $t$ was converted into a sinusoidal positional embedding, added to the corresponding frame's latents after a learnable projection~\cite{vaswani2023attention}. Each frame's local representation at spatial position ($h, w$), $\vz_{hwt}$, was then updated by attending to all frames at the same position $(\vz_{hw1}, \vz_{hw2}, \ldots,\vz_{hw(rT)})$ through a standard attention mechanism.

\newparagraph{Classifier-free Audio Guidance}
Classifier-free guidance~\cite{ho2022classifierfree} is a technique used to control the impact of input prompts on the generated outputs. We extended it to amplify audio guidance for improved synchronization. 
To accomplish this, we trained the model for both audio-conditioned and unconditioned generation, by randomly replacing $\va$ with a null audio embedding, $\va_\emptyset$, with a 20\% likelihood. $\va_\emptyset$ was computed by encoding an all-zero audio mel spectrogram via ImageBind. During inference, a factor $\eta\geq1$ controls the audio guidance:
\input{equations/audio_free_guidance}
 where $\eta$ guides the denoising process towards latents congruent with audio-conditioned generation and away from those of audio-unconditional generation. 
 

%% file: figures/arch_overview.tex
\begin{figure}[t!]
    \centering
    \includegraphics[width=0.85\linewidth]{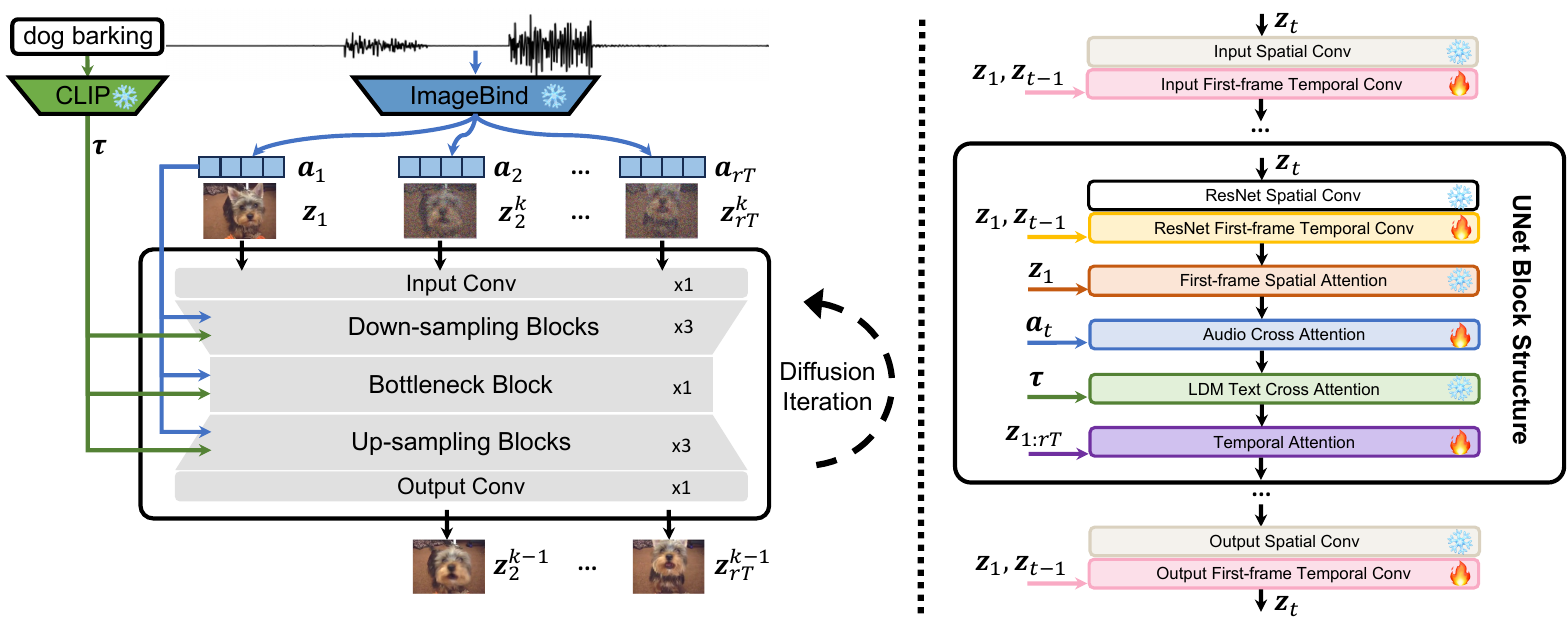}
    \caption{\method overview. \textit{Left}: We use ImageBind to encode audio into semantically aware time-dependent tokens $(\va_t)_{t=1}^{rT}$ and CLIP to encode the audio category into text embedding $\vtau$. In addition, the model receives the latent of the first frame $\vz_1$, and iteratively denoises noisy latents of the subsequent frames $\vz_{2:rT}^k$ via reverse diffusion. The denoising UNet, based on LDMs~\cite{robin2022ldm}, consists of a sequence of downsampling, bottleneck and up-sampling blocks, with structure detailed on the right.
    \textit{Right}: Anatomy of a UNet block for frame $\vz_t$. LDM's original spatial conv, spatial attention and text cross attention layers are frozen, while its spatial self-attention layers are adjusted to first-frame spatial attentions, cross-attending to $\vz_1$ instead. To learn video dynamics, we introduce temporal attention layers, and first-frame lookup temporal convolutions applied to input, output, and ResNet layers. We also train audio cross attentions for audio conditioning and synchronization. Trainable layers are marked with \fire.}
    \label{fig:arch_overview}
\end{figure}

%% file: equations/audio_free_guidance.tex
\begin{equation}
\vz_{2:rT}^{k-1} = (1-\eta) \cdot \epsilon_{\theta}(\vz_{2:rT}^k, k; \vz_1, \va_\emptyset, \vtau) + \eta \cdot \epsilon_{\theta}(\vz_{2:rT}^k, k; \vz_1, \va, \vtau)
\end{equation}

%% file: latex/5-experiments.tex
\section{Experiments}%
\label{sec:experiments}

\subsection{Implementation}%
\label{sec:implementation}

\newparagraph{Dataset} 
We conducted experiments on three datasets.~\textit{\db}: Our high-quality dataset curated from VGGSound~\cite{chen2020vggsound}, with 15 balanced categories, 1350 training videos, and 150 test videos. We also assessed our data curation pipeline by comparing it to models trained on Landscapes, TheGreatestHits, VGGSS and our AVSync-AC. \textit{Landscapes}~\cite{lee2022soundguided} is composed of 9 environmental sound classes. We followed the split in~\cite{ruan2022mmdiffusion,yariv2023tempotoken} with 900 clips for training and 100 for testing. Since Landscapes is full of ambient sounds without synchronized video motion, we mainly use it to evaluate visual quality. \textit{TheGreatestHits}~\cite{owens2016greatesthits} is an audio-video dataset recording humans probing environments with a drumstick, with 733 videos for training and 244 for testing. The videos are synchronized with audio at the moments of impact but contain lots of static moments. It also offers limited diversity, featuring a singular motion of probing. VGGSS and our AVSync-AC were described in \cref{sec:dataset}.

\newparagraph{Data Preprocess}
We sampled 2-second synchronized audio-video pairs for experiments.  Videos were sampled at 6 fps with 12 frames, and resized to 256$\times$256 on \db/Landscapes or 128$\times$256 on TheGreatestHits. Following ImageBind~\cite{girdhar2023imagebind}, audios were sampled at 16kHz and converted into 128-d spectrograms.

\newparagraph{Baselines} 
We first adopted a simple \textit{Static} baseline by repeating the input frame into a video, then compared it to several state-of-the-art works. 
(1) \textit{Semantic audio-to-video generation}~(CoDi~\cite{tang2023codi}): Image, text, and audio are encoded into a shared CLIP~\cite{radford2021CLIP} space and summed, then fused into a video diffusion model trained on large-scale datasets. 
(2) \textit{Image-to-video generation} (VideoCrafter \cite{chen2023videocrafter1}): A superior video diffusion model however without audio inputs. It animates images by fusing CLIP-encoded image and text features into the model via modality-dependent cross-attention layers.
(3) \textit{Synchronized audio-to-video generation}~(TPoS~\cite{jeong2023tpos}, AADiff~\cite{lee2023aadiff}, TempoToken~\cite{yariv2023tempotoken}): Audio is encoded into time-dependent segments and fused into frozen text-to-image~\cite{robin2022ldm} or text-to-video~\cite{wang2023modelscope} models. We re-implemented AADiff, and used TPoS and TempoToken's pretrained checkpoints on VGGSound and Landscapes. More details are provided in Suppl. Sec. 4.

\newparagraph{Training \& Evaluation} 
We adopted the pretrained Stable Diffusion-V1.5~\cite{robin2022ldm} as the diffusion model and ImageBind~\cite{girdhar2023imagebind} as the audio encoder. All models were trained using Adam optimizer with a batch size of 64 and a learning rate of 0.0001. Besides metrics in~\cref{section:dataset_metric}, we also provide results using metric in~\cite{difffoley2023} in Suppl. Sec. 3.4. We evaluated on 3 clips uniformly sampled from each video.

\subsection{Main Results}
\label{sec:main_results} 

\input{figures/benchmark_avsync15_landscapes_thegreatesthits}

\input{tables/main_result}
\newparagraph{Dataset Comparison}
\label{sec:dataset_comparison}
In \cref{tab:main_results}, the \textit{Static} baseline is a good indicator of dataset attributes. It has similar IA/IT scores compared to \textit{Groundtruth}, since \textit{Static} is composed of a subset of groundtruth frames, which are obviously semantically aligned. FVD of \textit{Static} on TheGreatestHits is relatively low, since the TheGreatestHits contains frequent static ground truth clips without any moving objects. More importantly, the RelSync and AlignSync of \textit{Static} gradually increase from \db{}, to TheGreatestHits, to Landscapes, with those on Landscapes even surpassing the ground truth. The fact that static videos perform well in terms of audio synchronization on TheGreatestHits and Landscapes datasets testify to the superiority of \db{} for audio-video synchronized generation. 


\newparagraph{Model Comparison}
\label{sec:model_comparison}
We compared to prior works on \db{} and Landscapes in \cref{tab:model_comparison_avsync15,tab:model_comparison_landscape,fig:benchmark_avsync15,fig:benchmark_landscapes}.
In \cref{tab:model_comparison_avsync15}, CoDi achieves inferior results on almost all metrics. 
TPoS(I+T+A) shows strong image quality (FID), but is worse in video quality (FVD) and synchronization (RelSync). TempoToken, on the other hand, is better at synchronization rather than visual quality, likely due to the lack of image input. AADiff is competitive on synchronization but extremely bad on image quality (FID). This is expected as AADiff adjusts each frame using audio amplitude, producing visual changes that highly correlate to audio changes temporally but may be overwhelmed by noises, as shown in \cref{fig:benchmark_avsync15}. On Landscapes, its similar FVD to \textit{Static} but abnormally higher AlignSync and RelSync also suggest that it only applied minor modifications to the input image due to lack of sound changes, as shown in \cref{fig:benchmark_landscapes}.
Without audio input, VideoCrafter performs poorly on synchronization. It also has difficulty faithfully adhering to the input frame, as in \cref{fig:benchmark_avsync15}. \method{} achieves the best animation results on almost all metrics. On Landscapes, \method{} also performs the best on FID and FVD, with other scores being similar to ground truth.

\newparagraph{User Study}
We invited 15 participants to compare 4 animation models with top overall performance (VideoCrafter, AADiff, I2VD, \method) on \db, based on 3 metrics in \cref{tab:main_results}. The 4 models generated videos conditioned on the same test examples (audio+image). Each test example was independently evaluated by 2 participants to select their most preferred generation (vote) on each metric. In total, we evaluated all 150 test examples on \db, collecting 150$\times$2=300 votes on each metric. \cref{tab:model_comparison_avsync15} shows votes each model received. 

\subsection{Ablation Studies}
\label{sec:ablation}

\newparagraph{Audio Conditioning}
In \cref{tab:main_results}, \method{} outperforms I2VD, especially on AlignSync and RelSync. \method{} did not improve RelSync on Landscapes probably due to the lack of synchronization cues on the dataset itself. These results show that audio condition enhances generation quality and synchronization.

\newparagraph{Audio Guidance}
\cref{tab:main_results} shows increasing the audio guidance factor $\eta$ from 1 to 4 improves FID, IA, and FVD on all three datasets. As expected, audio guidance also improved AlignSync and RelSync significantly on \db{} and TheGreatestHits, but not on the less synchronized Landscapes dataset. Prior works~\cite{sung2023soundtovisual,lee2023aadiff,jeong2023tpos} claimed that increasing audio amplitude can also lead to stronger visual effects. We compare this approach with our audio guidance in \cref{fig:audio_amplitude_cfg}. Audio guidance offers a better control mechanism than audio amplitude.

\newparagraph{First-frame~(FF) Lookups}
We validated FF Lookups by replacing them with standard temporal convolutions or spatial self-attention in \cref{tab:model_ablation}.

\newparagraph{Data Curation}
We compared to \method{} trained on random subsets from VGGSS and AVSync-AC with equal data scale and balanced categories in \cref{tab:dataset_ablation}.

\subsection{Applications and Extensions}
\label{sec:application}

\input{tables/model_dataset_ablations.tex}
\input{figures/audio_amplitude_cfg}
\input{figures/related_sound_control}

\newparagraph{Animate Generated Images} When lacking an image as input, we can use existing image generators to generate the image, which \method{} can also animate. \cref{fig:generated_control} shows animations on images generated by StableDiffusion-V1.5~\cite{robin2022ldm}.

\newparagraph{Animate Contents from Internet} \method{} can also generalize well to unseen images and audio, as shown in \cref{fig:wild_control}.

\newparagraph{Control Animations with (Un)Related Audios} We can control the motion of an image to follow desired audio, e.g., animate a baby to not only cry but also bark like a dog or roar like a lion, as seen in \cref{fig:related_sound_control}. When there is no object related to the audio, the animations do not demonstrate corresponding motion.

\newparagraph{Animate Target Objects with Audios} When multiple objects exist in the image, a scenario not existing in training data, we can still use audios to only animate the related target object, as shown in \cref{fig:multi_object_control}.

%% file: figures/benchmark_avsync15_landscapes_thegreatesthits.tex
\begin{figure}[t!]
    \begin{subfigure}[b]{\linewidth}
        \centering
        \includegraphics[width=\linewidth]{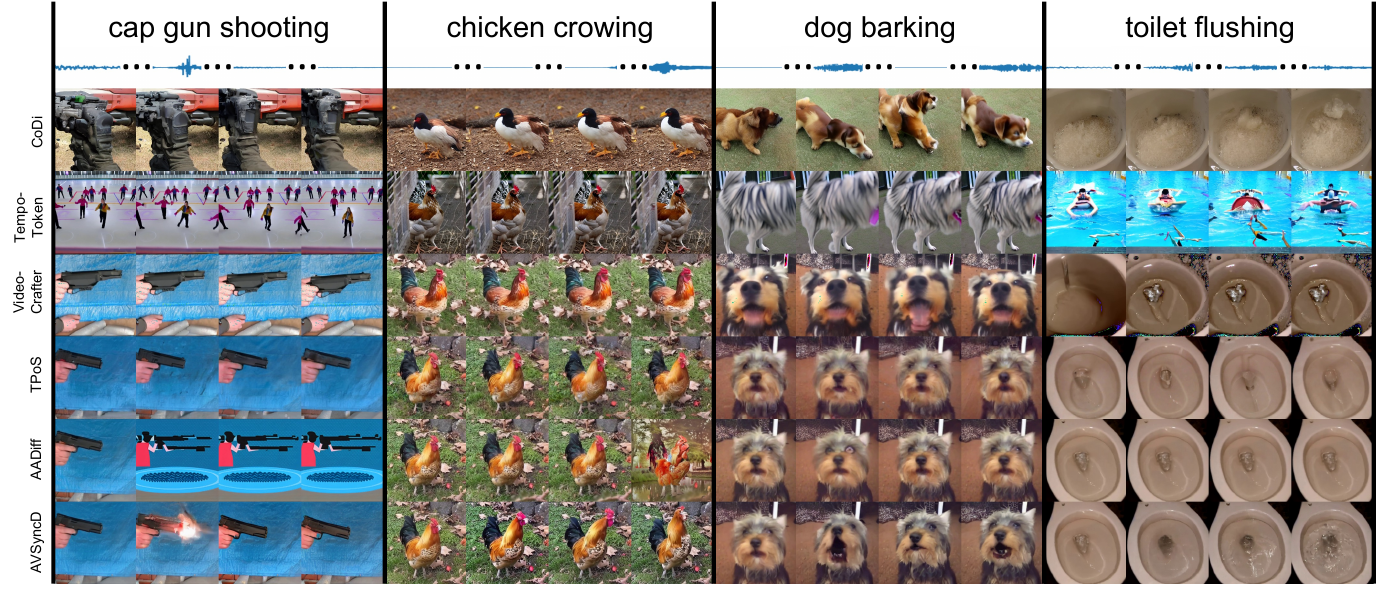}
        \caption{Comparison on \db. \method produces synchronized dynamic motions, such as muzzle flash, chicken and dog's head poses, and swirling in toilet.}
        \label{fig:benchmark_avsync15}
    \end{subfigure}
    \hfill
    \begin{subfigure}[b]{0.58\linewidth}
        \centering
        \includegraphics[width=\linewidth]{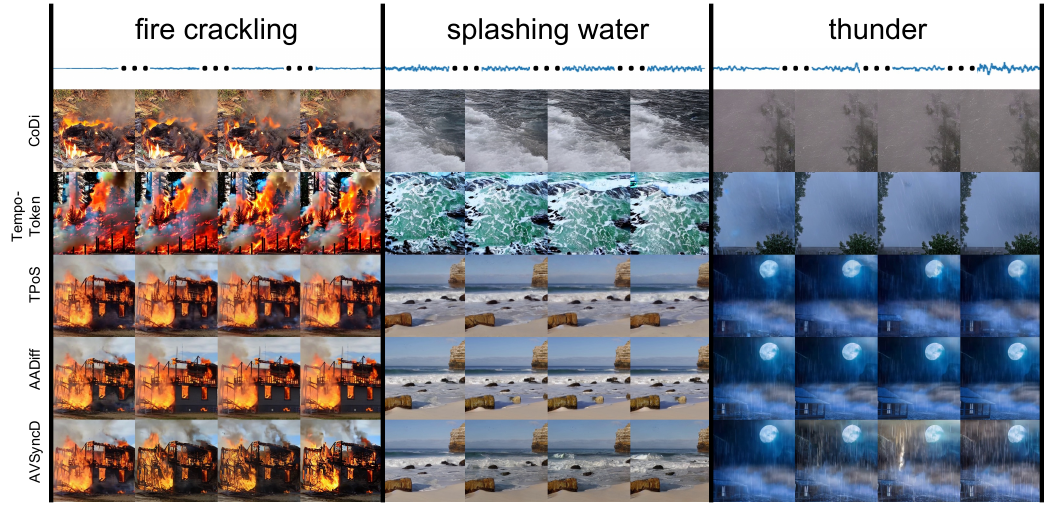}
        \caption{Comparison on Landscapes. Without synchronization cues in audio, \method still produces evolving visual effects and semantics, such as flames, waves, and lightning.}
        \label{fig:benchmark_landscapes}
     \end{subfigure}
     \hfill
     \begin{subfigure}[b]{0.41\linewidth}
         \centering
         \includegraphics[width=\linewidth]{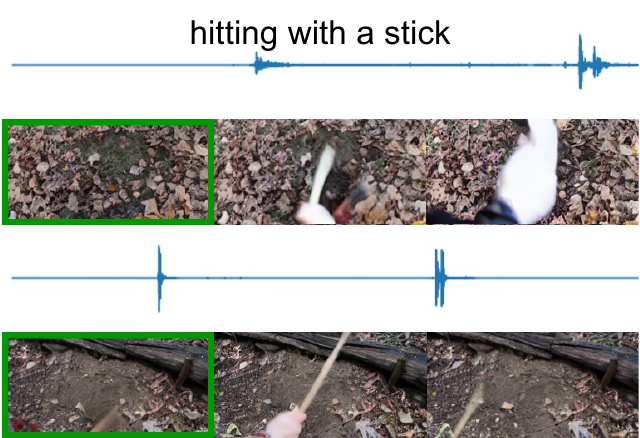}
         \caption{Generation by \method on TheGreatestHits. \method generates \textit{hitting} actions at appropriate moments.}
         \label{fig:qualitative_view_thegreatesthits}
     \end{subfigure}
     \caption{ Qualitative results on three datasets.} 
     \label{fig:benchmark}
\end{figure}

%% file: tables/main_result.tex
\begin{table}[t!]
\caption{ Quantitative results. User study shows votes on 3 metrics: image quality, frame consistency, and synchronization. Inputs are combinations of image, text, audio.}
\centering

\noindent
\begin{subtable}{1.0\linewidth}
    \resizebox{\linewidth}{!}{
    \centering
    \begin{tabular}{l l c c c c c c c c c c c}
    \toprule
        \bf\multirow{2}{*}{Input} & \bf\multirow{2}{*}{Model} & \bf\multirow{2}{*}{FID$\downarrow$} & \bf\multirow{2}{*}{IA$\uparrow$} & \bf\multirow{2}{*}{IT$\uparrow$} & \bf\multirow{2}{*}{FVD$\downarrow$} & \bf\multirow{2}{*}{AlignSync$\uparrow$} & \bf\multirow{2}{*}{RelSync$\uparrow$} & \multicolumn{3}{c}{\bf User Study}\\
        & & & & & & & & \bf IQ$\uparrow$ & \bf FC$\uparrow$ & \bf Sync$\uparrow$ \\
         \midrule
         \multirow{2}{*}{T+A} 
         & TPoS~\cite{jeong2023tpos}                  & $13.5$ & $23.38$ & $24.83$ & $2671.0$ & $19.52$ & $42.50$ & - & - & - \\
         & TempoToken~\cite{yariv2023tempotoken}      & $12.2$ & $18.84$ & $17.45$ & $4466.4$ & $19.74$ & $44.05$ & - & - & - \\
         \midrule
         \multirow{2}{*}{I+T}
         & VideoCrafter~\cite{chen2023videocrafter1}  & $11.8$ & - & $29.87$ & $840.7$ & $21.28$ & $43.16$ & $38$ & $20$ & $12$ \\
         & I2VD                                       & $12.1$ & - & $30.35$ & $398.2$ & $21.80$ & $43.92$ & $62$ & $90$ & $91$ \\
         \midrule
         \multirow{6}{*}{I+T+A}
         & CoDi~\cite{tang2023codi}                   & $14.5$ & $28.15$ & $23.42$ & $1522.6$ & $19.54$ & $41.51$ & - & - & - \\
         & TPoS~\cite{jeong2023tpos}                  & $11.9$ & $38.36$ & $\bf30.73$ & $1227.8$ & $19.67$ & $39.62$ & - & - & - \\
         & AADiff~\cite{lee2023aadiff}                & $18.8$ & $34.23$ & $28.97$ & $978.0$ & $22.11$ & $45.48$ & $37$ & $4$ & $5$\\
         & \method $\eta=1$                           & $12.1$ & $38.36$ & $30.34$ & $382.7$ & $22.25$ & $44.81$ & - & - & - \\
         & \method $\eta=4$                           & $\bf11.7$ & $\bf38.53$ & $30.45$ & $\bf349.1$ & $22.62$ & $45.52$ & $\bf163$ & $\bf186$ & $\bf192$ \\
         & \method $\eta=8$                           & $\bf11.7$ & $37.99$ & $30.27$ & $420.7$ & $\bf22.74$ & $\bf45.88$ & - & - & - \\
         \midrule
         \multicolumn{2}{c}{Static}                                     & - & $39.76$ & $30.39$ & $1220.4$ & $21.83$ & $43.66$ \\
         \multicolumn{2}{c}{Groundtruth}                                & - & $40.06$ & $30.31$ & - & $25.04$ & $50.00$ \\
    \bottomrule
    \end{tabular}}
    \caption{Performance on \db.}
    \label{tab:model_comparison_avsync15}
\end{subtable}
\hfill
\begin{subtable}{1.0\linewidth}
    \setlength{\tabcolsep}{12pt} 
    \resizebox{\linewidth}{!}{
    \centering
    \begin{tabular}{l l c c c c c c c c}
    \toprule
        \bf Input & \bf Model & \bf FID$\downarrow$ & \bf IA$\uparrow$ & \bf IT$\uparrow$ & \bf FVD$\downarrow$ & \bf AignSync$\uparrow$ & \bf RelSync$\uparrow$ \\
         \midrule
         \multirow{2}{*}{T+A} 
         & TPoS~\cite{jeong2023tpos}               & $16.5$ & $15.61$ & $\bf26.70$ & $2081.3$ & $23.12$ & $48.15$ \\
         & TempoToken~\cite{yariv2023tempotoken}   & $16.4$ & $22.58$ & $22.87$ & $2480.0$ & $24.21$ & $48.65$ \\
         \midrule
         \multirow{1}{*}{I+T}
         & I2VD                                    & $16.7$ & - & $22.56$ & $539.5$ & $24.74$ & $49.89$ \\
         \midrule
         \multirow{5}{*}{I+T+A}
         & CoDi~\cite{tang2023codi}                & $20.5$ & $22.63$ & $24.23$ & $982.9$ & $22.63$ & $45.48$ \\
         & TPoS~\cite{jeong2023tpos}               & $\bf16.2$ & $\bf23.52$ & $23.20$ & $789.6$ & $23.51$ & $47.05$ \\
         & AADiff~\cite{lee2023aadiff}             & $70.7$ & $22.07$ & $22.92$ & $1186.3$ & $\bf26.77$ & $\bf53.93$ \\
         & \method $\eta=1$                        & $16.5$ & $22.29$ & $22.81$ & $463.1$ & $24.81$ & $49.96$ \\
         & \method $\eta=4$                        & $\bf16.2$ & $22.49$ & $22.79$ & $\bf415.2$ & $24.82$ & $49.93$ \\
         \midrule
         \multicolumn{2}{c}{Static}                & - & $23.60$ & $22.21$ & $1177.5$ & $25.79$ & $51.59$ \\
         \multicolumn{2}{c}{Groundtruth}           & - & $23.65$ & $22.08$ & - & $25.01$ & $50.00$ \\
    \bottomrule
    \end{tabular}}
    \caption{Performance on Landscapes.}
    \label{tab:model_comparison_landscape}
\end{subtable}
\centering
\begin{subtable}{1.0\linewidth}
    \setlength{\tabcolsep}{12pt} 
    \resizebox{\linewidth}{!}{
    \begin{tabular}{l l c c c c c c c c}
    \toprule
         \bf Input & \bf Model & \bf FID$\downarrow$ & \bf IA$\uparrow$ & \bf IT$\uparrow$ & \bf FVD$\downarrow$ & \bf AignSync$\uparrow$ & \bf RelSync$\uparrow$ \\
         \midrule
         \multirow{1}{*}{I+T}
         & I2VD                   & $9.1$ & - & $\bf13.42$ & $425.0$ & $22.05$ & $44.58$ \\
         \midrule
         \multirow{2}{*}{I+T+A}
         & \method $\eta=1$       & $9.0$ & $11.85$ & $13.18$ & $313.5$ & $22.59$ & $45.52$ \\
         & \method $\eta=4$       & $\bf8.7$ & $\bf12.07$ & $13.31$ & $\bf249.3$ & $\bf22.83$ & $\bf45.95$ \\
         \midrule
         \multicolumn{2}{c}{Static}                 & - & $13.33$ & $16.56$ & $348.9$ & $24.36$ & $48.73$ \\
         \multicolumn{2}{c}{Groundtruth}            & -  & $13.52$ & $16.49$ & - & $25.02$ & $50.00$  \\
    \bottomrule
    \end{tabular}}
    \caption{Performance on TheGreatestHits.}
    \label{tab:model_thegreatesthits}
\end{subtable}
\label{tab:main_results}
\end{table}

%% file: tables/model_dataset_ablations.tex
\setlength{\tabcolsep}{1.5pt} 
\begin{table}[t]
\centering
\small
\caption{Effect of (a) first-frame lookups (b) data curation, evaluated on \db.}
    
\noindent
\begin{subtable}{0.48\linewidth}
    \setlength{\tabcolsep}{4pt} 
    \resizebox{\linewidth}{!}{
    \begin{tabular}{c c c c c}
        \toprule
        \bf FF-Conv & \bf FF-Attn & \bf FID$\downarrow$ & \bf FVD$\downarrow$ & \bf AignSync$\uparrow$ \\
        \midrule
        \xmark & \xmark & $11.8$ & $383.3$ & $22.19$ \\
        \cmark & \xmark & $\bf11.6$ &$347.4$ & $22.24$ \\
        \cmark & \cmark & $11.8$ & $\bf325.6$ & $\bf22.33$ \\
        \bottomrule
    \end{tabular}}
    \caption{}
    \label{tab:model_ablation}
\end{subtable}
\hfill
\begin{subtable}{0.48\linewidth}
    \setlength{\tabcolsep}{4pt} 
    \resizebox{\linewidth}{!}{
    \begin{tabular}{c c c c c c}
        \toprule
        \bf Dataset & \bf AC & \bf MC & \bf FID$\downarrow$ & \bf FVD$\downarrow$ & \bf AignSync$\uparrow$ \\
        \midrule
        VGGSS           & \xmark & \xmark & $12.9$ & $1307.9$ & $21.50$ \\
        AVSync-AC       & \cmark & \xmark & $12.0$ & $428.8$ & $22.09$ \\
        \db             & \cmark & \cmark & $\bf11.8$ & $\bf325.6$ & $\bf22.33$ \\
        \bottomrule
    \end{tabular}}
    \caption{}
    \label{tab:dataset_ablation}
\end{subtable}

\end{table}

%% file: figures/audio_amplitude_cfg.tex
\begin{figure}[t!]
    \begin{subfigure}[b]{0.54\linewidth}
        \centering
        \includegraphics[width=\linewidth]{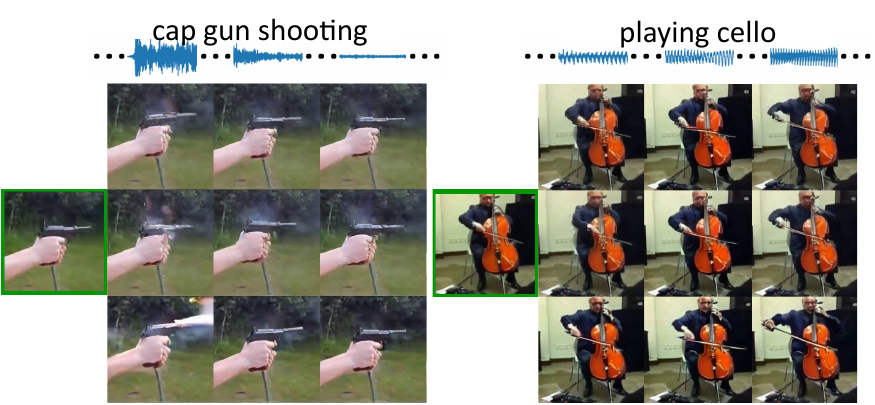}
        \caption{}
        \label{fig:audio_amplitude_cfg}
    \end{subfigure}
    \hfill
    \begin{subfigure}[b]{0.215\linewidth}
        \centering
        \includegraphics[width=\linewidth]{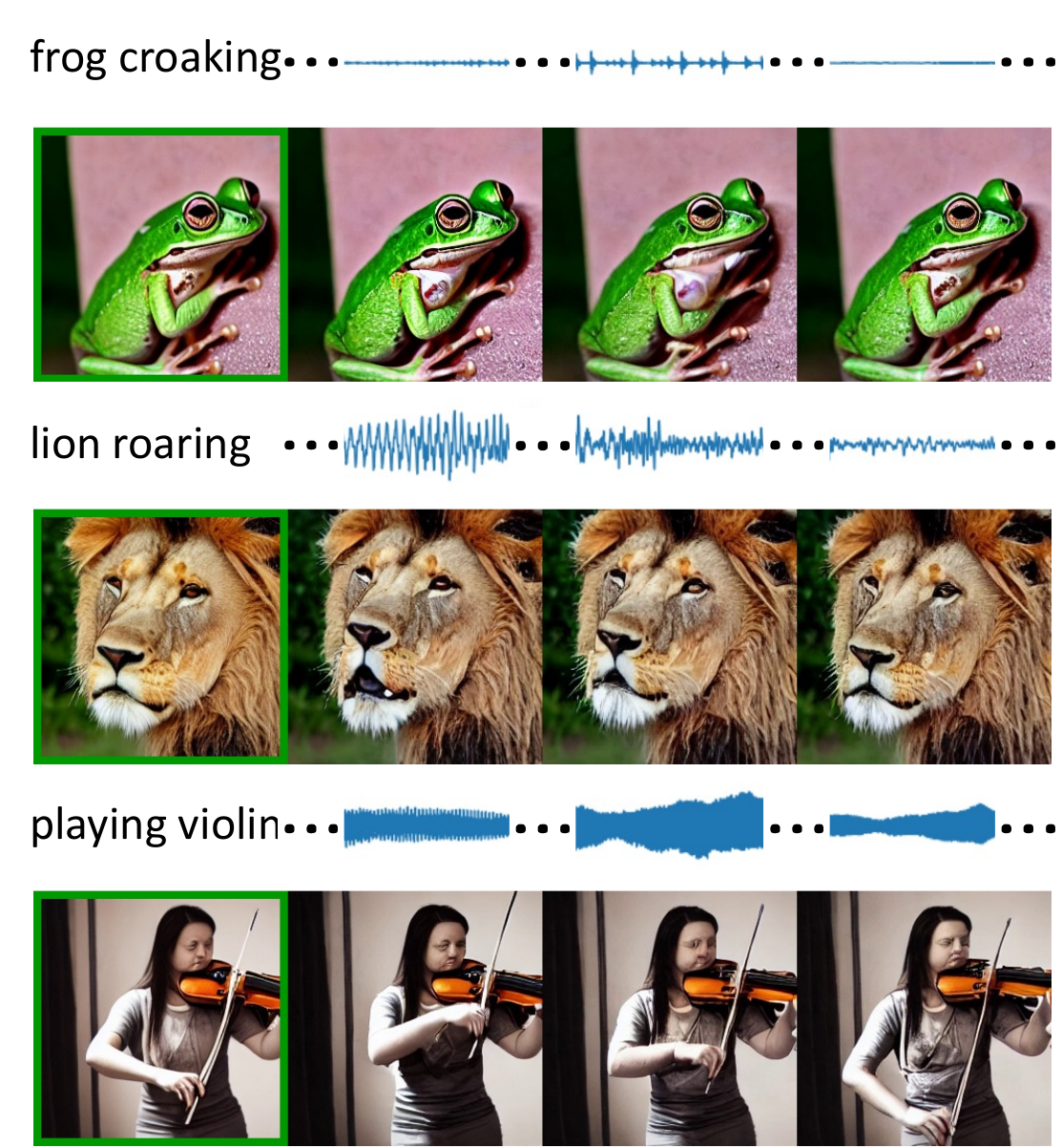}
        \caption{}
        \label{fig:generated_control}
    \end{subfigure}
    \hfill
    \begin{subfigure}[b]{0.215\linewidth}
        \centering
        \includegraphics[width=\linewidth]{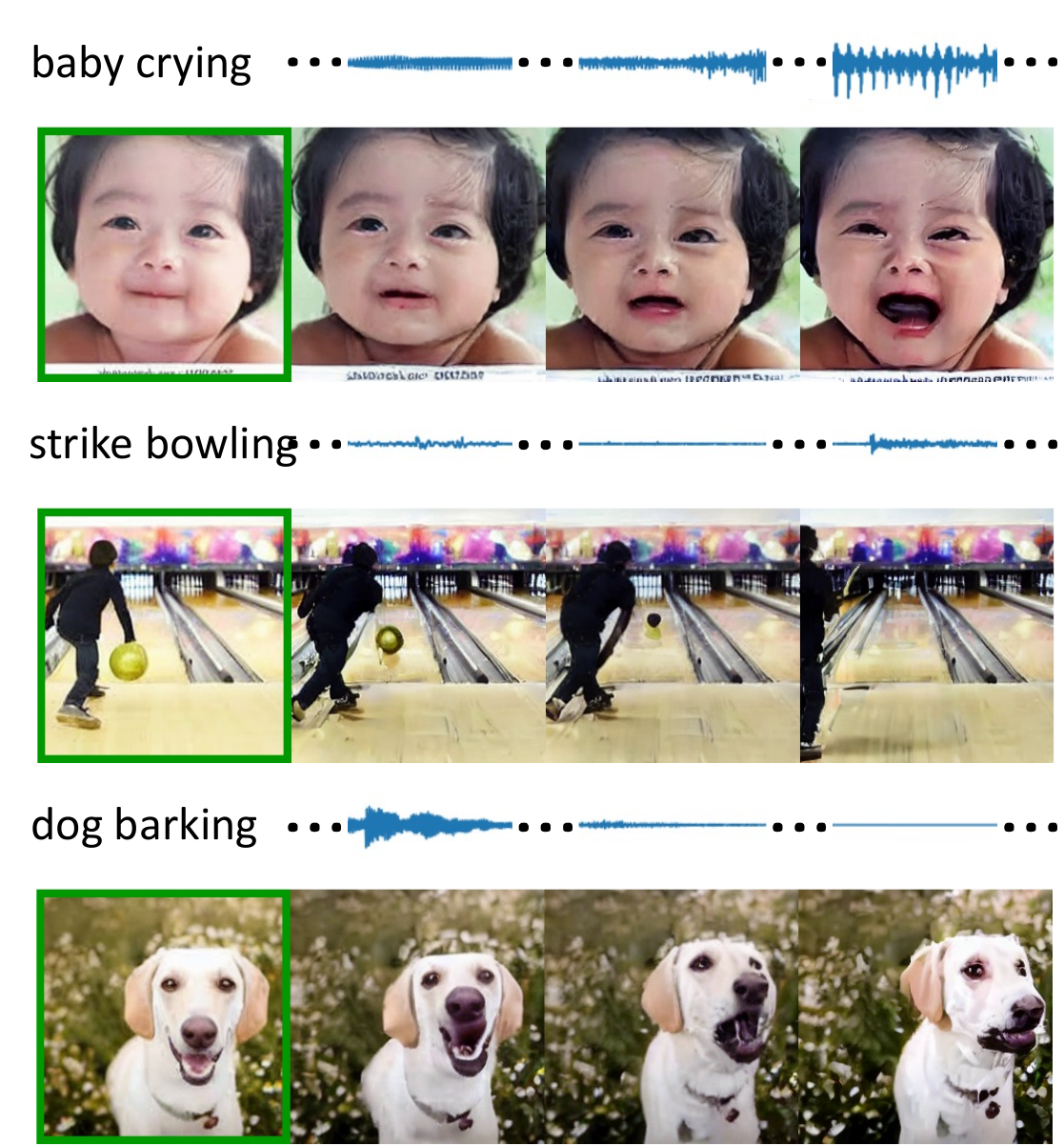}
        \caption{}
        \label{fig:wild_control}
    \end{subfigure}
    \caption{(a): Effects of audio amplitude vs. classifier-free audio guidance. \textit{top}: original audio with $\eta=1$; \textit{mid}: $100\times$ amplified audio with $\eta=1$; \textit{bottom}: original audio with $\eta=8$. (b): Animate generated images. (c): Animation with internet images and audios.}
\end{figure}

%% file: figures/related_sound_control.tex
\begin{figure}[t!]
    \begin{subfigure}[b]{\linewidth}
        \centering
        \includegraphics[width=\linewidth]{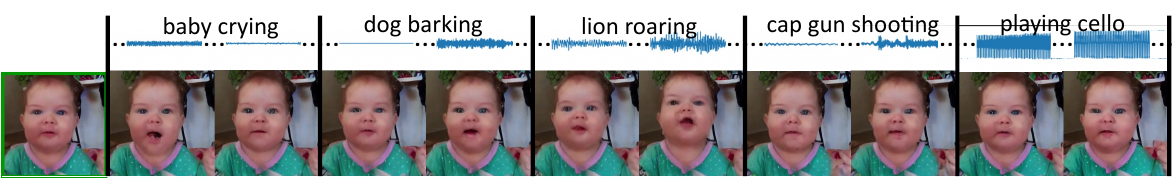}
        \caption{Controllable animation with (un)related audios.}
        \label{fig:related_sound_control}
    \end{subfigure}
    \hfill
    \begin{subfigure}[b]{\linewidth}
        \centering
        \includegraphics[width=\linewidth]{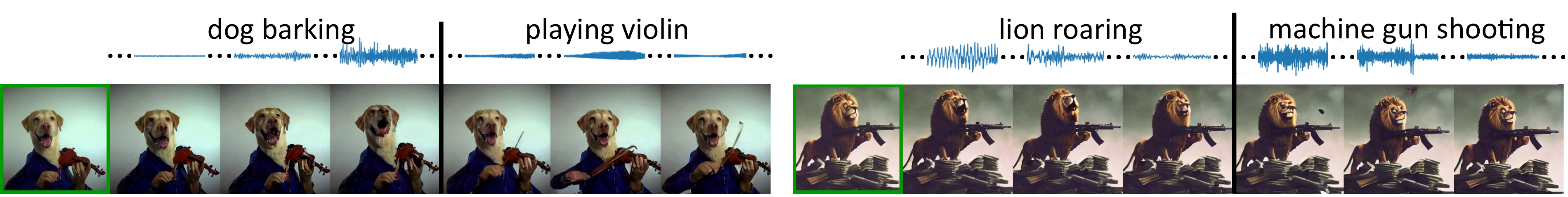}
        \caption{Animate target objects on internet images with audios.}
        \label{fig:multi_object_control}
    \end{subfigure}
    \caption{Controllable image animation with audios. Key frames are visualized. }
\end{figure}

%% file: latex/6-conclusion.tex
\section{Conclusion}\label{sec:conclusion}
We tackled the under-explored Audio-Synchronized Visual Animation task, with an emphasis on generating videos with audio-synchronized dynamics. We contributed the high-quality \db{} benchmark via careful data curation and proposed the \method{} model to animate images with realistic motions. Due to the scale of \db, our work cannot generalize to all audio classes in the world, which requires several orders of magnitude larger datasets. However, we hope our research inspires further work in this direction.

%% file: main.bbl
\begin{thebibliography}{10}
\providecommand{\url}[1]{\texttt{#1}}
\providecommand{\urlprefix}{URL }
\providecommand{\doi}[1]{https://doi.org/#1}

\bibitem{Bain21webvid10m}
Bain, M., Nagrani, A., Varol, G., Zisserman, A.: Frozen in time: A joint video and image encoder for end-to-end retrieval. In: IEEE International Conference on Computer Vision (2021)

\bibitem{blattmann2023videoldm}
Blattmann, A., Rombach, R., Ling, H., Dockhorn, T., Kim, S.W., Fidler, S., Kreis, K.: Align your latents: High-resolution video synthesis with latent diffusion models. In: IEEE Conference on Computer Vision and Pattern Recognition ({CVPR}) (2023)

\bibitem{brock2019biggan}
Brock, A., Donahue, J., Simonyan, K.: Large scale gan training for high fidelity natural image synthesis. In: ICLR (2019)

\bibitem{pyscenedetect}
Castellano, B.: Pyscenedetect. \url{https://www.scenedetect.com/}

\bibitem{chen2023videocrafter1}
Chen, H., Xia, M., He, Y., Zhang, Y., Cun, X., Yang, S., Xing, J., Liu, Y., Chen, Q., Wang, X., Weng, C., Shan, Y.: Videocrafter1: Open diffusion models for high-quality video generation (2023)

\bibitem{chen2021vggsoundsync}
Chen, H., Xie, W., Afouras, T., Nagrani, A., Vedaldi, A., Zisserman, A.: Audio-visual synchronization in the wild. In: Proceedings of the British Machine Vision Conference (BMVC) (2021)

\bibitem{chen2020vggsound}
Chen, H., Xie, W., Vedaldi, A., Zisserman, A.: Vggsound: A large-scale audio-visual dataset. In: ICASSP (2020)

\bibitem{Chen2022beats}
Chen, S., Wu, Y., Wang, C., Liu, S., Tompkins, D., Chen, Z., Wei, F.: Beats: Audio pre-training with acoustic tokenizers. In: ICML (2023)

\bibitem{lipreading}
Chung, J.S., Zisserman, A.: Lip reading in the wild. In: ACCV (2016)

\bibitem{outoftime}
Chung, J.S., Zisserman, A.: Out of time: automated lip sync in the wild. In: ACCV Workshop (2016)

\bibitem{Chung2019perfectmatch}
Chung, S.W., Chung, J.S., Kang, H.G.: Perfect match: Improved cross-modal embeddings for audio-visual synchronisation. In: ICASSP 2019 - 2019 IEEE International Conference on Acoustics, Speech and Signal Processing (ICASSP) (2019)

\bibitem{dhariwal2021diffusionbeatgan}
Dhariwal, P., Nichol, A.: Diffusion models beat gans on image synthesis. In: NeurIPS (2021)

\bibitem{audioset}
Gemmeke, J.F., Ellis, D.P.W., Freedman, D., Jansen, A., Lawrence, W., Moore, R.C., Plakal, M., Ritter, M.: Audio set: An ontology and human-labeled dataset for audio events. In: Proc. IEEE ICASSP 2017 (2017)

\bibitem{girdhar2023imagebind}
Girdhar, R., El-Nouby, A., Liu, Z., Singh, M., Alwala, K.V., Joulin, A., Misra, I.: Imagebind: One embedding space to bind them all. In: CVPR (2023)

\bibitem{hertz2022prompt}
Hertz, A., Mokady, R., Tenenbaum, J., Aberman, K., Pritch, Y., Cohen-Or, D.: Prompt-to-prompt image editing with cross attention control. In: ICLR (2023)

\bibitem{fid}
Heusel, M., Ramsauer, H., Unterthiner, T., Nessler, B., Hochreiter, S.: Gans trained by a two time-scale update rule converge to a local nash equilibrium. In: Advances in Neural Information Processing Systems (2017)

\bibitem{ho2020ddpm}
Ho, J., Jain, A., Abbeel, P.: Denoising diffusion probabilistic models. In: NeurIPS (2020)

\bibitem{ho2022classifierfree}
Ho, J., Salimans, T.: Classifier-free diffusion guidance. In: NeurIPS Workshop on Deep Generative Models and Downstream Applications (2022)

\bibitem{sparse2022iashin}
Iashin, V., Xie, W., Rahtu, E., Zisserman, A.: Sparse in space and time: Audio-visual synchronisation with trainable selectors. In: British Machine Vision Conference (BMVC) (2022)

\bibitem{jeong2023tpos}
Jeong, Y., Ryoo, W., Lee, S., Seo, D., Byeon, W., Kim, S., Kim, J.: The power of sound (tpos): Audio reactive video generation with stable diffusion. In: Proceedings of the IEEE/CVF International Conference on Computer Vision. pp. 7822--7832 (2023)

\bibitem{text2video-zero}
Khachatryan, L., Movsisyan, A., Tadevosyan, V., Henschel, R., Wang, Z., Navasardyan, S., Shi, H.: Text2video-zero: Text-to-image diffusion models are zero-shot video generators. In: ICCV (2023)

\bibitem{lee2022soundguided}
Lee, S.H., Oh, G., Byeon, W., Bae, J., Kim, C., Ryoo, W.J., Yoon, S.H., Kim, J., Kim, S.: Sound-guided semantic video generation. In: ECCV (2022)

\bibitem{lee2023aadiff}
Lee, S., Kong, C., Jeon, D., Kwak, N.: Aadiff: Audio-aligned video synthesis with text-to-image diffusion. In: CVPR Workshop on Content Generation (2023)

\bibitem{li2021aist++}
Li, R., Yang, S., Ross, D.A., Kanazawa, A.: Learn to dance with aist++: Music conditioned 3d dance generation. In: ICCV (2021)

\bibitem{li2023gligen}
Li, Y., Liu, H., Wu, Q., Mu, F., Yang, J., Gao, J., Li, C., Lee, Y.J.: Gligen: Open-set grounded text-to-image generation. In: CVPR (2023)

\bibitem{difffoley2023}
Luo, S., Yan, C., Hu, C., Zhao, H.: Diff-foley: Synchronized video-to-audio synthesis with latent diffusion models. In: NeurIPS (2023)

\bibitem{mokady2022null}
Mokady, R., Hertz, A., Aberman, K., Pritch, Y., Cohen-Or, D.: Null-text inversion for editing real images using guided diffusion models. In: CVPR (2023)

\bibitem{ng2024audio2photoreal}
Ng, E., Romero, J., Bagautdinov, T., Bai, S., Darrell, T., Kanazawa, A., Richard, A.: From audio to photoreal embodiment: Synthesizing humans in conversations. In: ArXiv (2024)

\bibitem{owens2016greatesthits}
Owens, A., Isola, P., McDermott, J., Torralba, A., Adelson, E.H., Freeman, W.T.: Visually indicated sounds. In: CVPR (2016)

\bibitem{park2022synctalkface}
Park, S.J., Kim, M., Hong, J., Choi, J., Ro, Y.M.: Synctalkface: Talking face generation with precise lip-syncing via audio-lip memory. In: AAAI Conference on Artificial Intelligence (AAAI) (2022)

\bibitem{morgado2021_robust_xid}
Pedro~Morgado, Ishan~Misra, N.V.: Robust audio-visual instance discrimination. In: Computer Vision and Pattern Recognition (CVPR), IEEE/CVF Conf. on (2021)

\bibitem{radford2021CLIP}
Radford, A., Kim, J.W., Hallacy, C., Ramesh, A., Goh, G., Agarwal, S., Sastry, G., Askell, A., Mishkin, P., Clark, J., Krueger, G., Sutskever, I.: Learning transferable visual models from natural language supervision. In: ICML (2021)

\bibitem{ramesh2022dalle2}
Ramesh, A., Dhariwal, P., Nichol, A., Chu, C., Chen, M.: Hierarchical text-conditional image generation with clip latents. In: arXiv (2022)

\bibitem{robin2022ldm}
Rombach, R., Blattmann, A., Lorenz, D., Esser, P., Ommer, B.: High-resolution image synthesis with latent diffusion models. In: CVPR (2022)

\bibitem{ruan2022mmdiffusion}
Ruan, L., Ma, Y., Yang, H., He, H., Liu, B., Fu, J., Yuan, N.J., Jin, Q., Guo, B.: Mm-diffusion: Learning multi-modal diffusion models for joint audio and video generation. In: CVPR (2023)

\bibitem{schuhmann2022laion5b}
Schuhmann, C., Beaumont, R., Vencu, R., Gordon, C., Wightman, R., Cherti, M., Coombes, T., Katta, A., Mullis, C., Wortsman, M., Schramowski, P., Kundurthy, S., Crowson, K., Schmidt, L., Kaczmarczyk, R., Jitsev, J.: Laion-5b: An open large-scale dataset for training next generation image-text models. In: NeurIPS (2022)

\bibitem{singer2022makeavideo}
Singer, U., Polyak, A., Hayes, T., Yin, X., An, J., Zhang, S., Hu, Q., Yang, H., Ashual, O., Gafni, O., Parikh, D., Gupta, S., Taigman, Y.: Make-a-video: Text-to-video generation without text-video data (2022)

\bibitem{song2022ddim}
Song, J., Meng, C., Ermon, S.: Denoising diffusion implicit models. In: ICLR (2021)

\bibitem{sung2023soundtovisual}
Sung-Bin, K., Senocak, A., Ha, H., Owens, A., Oh, T.H.: Sound to visual scene generation by audio-to-visual latent alignment. In: IEEE/CVF Conference on Computer Vision and Pattern Recognition (CVPR) (2023)

\bibitem{tang2023codi}
Tang, Z., Yang, Z., Zhu, C., Zeng, M., Bansal, M.: Any-to-any generation via composable diffusion. In: Thirty-seventh Conference on Neural Information Processing Systems (2023), \url{https://openreview.net/forum?id=2EDqbSCnmF}

\bibitem{aist-dance-db}
Tsuchida, S., Fukayama, S., Hamasaki, M., Goto, M.: Aist dance video database: Multi-genre, multi-dancer, and multi-camera database for dance information processing. In: Proceedings of the 20th International Society for Music Information Retrieval Conference, {ISMIR} 2019. Delft, Netherlands (Nov 2019)

\bibitem{fvd}
Unterthiner, T., van Steenkiste, S., Kurach, K., Marinier, R., Michalski, M., Gelly, S.: Towards accurate generative models of video: A new metric \& challenges. In: arXiv (2019)

\bibitem{vaswani2023attention}
Vaswani, A., Shazeer, N., Parmar, N., Uszkoreit, J., Jones, L., Gomez, A.N., Kaiser, L., Polosukhin, I.: Attention is all you need. In: NeurIPS (2017)

\bibitem{wang2023modelscope}
Wang, J., Yuan, H., Chen, D., Zhang, Y., Wang, X., Zhang, S.: Modelscope text-to-video technical report (2023)

\bibitem{vid2vid-zero}
Wang, W., Xie, k., Liu, Z., Chen, H., Cao, Y., Wang, X., Shen, C.: Zero-shot video editing using off-the-shelf image diffusion models. arXiv preprint arXiv:2303.17599  (2023)

\bibitem{wu2023lamp}
Wu, R., Chen, L., Yang, T., Guo, C., Li, C., Zhang, X.: Lamp: Learn a motion pattern by few-shot tuning a text-to-image diffusion model. arXiv preprint arXiv:2310.10769  (2023)

\bibitem{xu2021videoclip}
Xu, H., Ghosh, G., Huang, P.Y., Okhonko, D., Aghajanyan, A., Metze, F., Zettlemoyer, L., Feichtenhofer, C.: Videoclip: Contrastive pre-training for zero-shot video-text understanding. In: EMNLP (2021)

\bibitem{yariv2023tempotoken}
Yariv, G., Gat, I., Benaim, S., Wolf, L., Schwartz, I., Adi, Y.: Diverse and aligned audio-to-video generation via text-to-video model adaptation (2023)

\bibitem{ye2023geneface}
Ye, Z., Jiang, Z., Ren, Y., Liu, J., He, J., Zhao, Z.: Geneface: Generalized and high-fidelity audio-driven 3d talking face synthesis. In: ICLR (2023)

\bibitem{zhang2023adding}
Zhang, L., Rao, A., Agrawala, M.: Adding conditional control to text-to-image diffusion models. In: IEEE International Conference on Computer Vision (ICCV) (2023)

\bibitem{zhou2019talking}
Zhou, H., Liu, Y., Liu, Z., Luo, P., Wang, X.: Talking face generation by adversarially disentangled audio-visual representation. In: AAAI Conference on Artificial Intelligence (AAAI) (2019)

\bibitem{Zhou2021Pose}
Zhou, H., Sun, Y., Wu, W., Loy, C.C., Wang, X., Liu, Z.: Pose-controllable talking face generation by implicitly modularized audio-visual representation. In: Proceedings of the IEEE Conference on Computer Vision and Pattern Recognition (CVPR) (2021)

\end{thebibliography}
